\documentclass[runningheads]{llncs}
\usepackage[ruled,vlined,linesnumbered]{algorithm2e}
\usepackage{graphicx}
\usepackage{tabularx} 
\usepackage{amsmath}
\usepackage{amssymb}
\usepackage{multirow}
\usepackage{subfigure}
\usepackage{xcolor}
\usepackage{url}

\begin{document}
\title{Multiple Meta-model Quantifying for Medical Visual Question Answering}

\titlerunning{Multiple Meta-model Quantifying for Medical Visual Question Answering}
%
\author{Tuong Do\inst{1}\and
Binh X. Nguyen\inst{1}\and
Erman Tjiputra\inst{1}\and
Minh Tran\inst{1}\and\\
Quang D. Tran\inst{1}\and
Anh Nguyen\inst{2}}
\institute{AIOZ, Singapore \\ \email{\{tuong.khanh-long.do,binh.xuan.nguyen,erman.tjiputra,\\minh.quang.tran,quang.tran\}@aioz.io} \and University of Liverpool, UK \\
\email{anh.nguyen@liverpool.ac.uk}}
\maketitle              
\begin{abstract}
Transfer learning is an important step to extract meaningful features and overcome the data limitation in the medical Visual Question Answering (VQA) task. However, most of the existing medical VQA methods rely on external data for transfer learning, while the meta-data within the dataset is not fully utilized. In this paper, we present a new multiple meta-model quantifying method that effectively learns meta-annotation and leverages meaningful features to the medical VQA task. Our proposed method is designed to increase meta-data by auto-annotation, deal with noisy labels, and output meta-models which provide robust features for medical VQA tasks. Extensively experimental results on two public medical VQA datasets show that our approach achieves superior accuracy in comparison with other state-of-the-art methods, while does not require external data to train meta-models. 
Source code available at: \url{https://github.com/aioz-ai/MICCAI21_MMQ}.
\end{abstract}

\section{Introduction}
A medical Visual Question Answering (VQA) system can provide meaningful references for both doctors and patients during the treatment process. 
Extracting image features is one of the most important steps in a medical VQA framework which outputs essential information to predict answers.
Transfer learning, in which the pretrained deep learning models~\cite{Simonyan2015VGG,He2016ResNet,nguyen2020end,baurudepth21,huang2020tracking} that are trained on the large scale labeled dataset such as ImageNet~\cite{russakovsky2015imagenet}, is a popular way to initialize the feature extraction process. However, due to the difference in visual concepts between ImageNet images and medical images, finetuning process is not sufficient \cite{nguyen2019overcoming}. 
Recently, Model Agnostic Meta-Learning \cite{finn2017model} (MAML) has been introduced to overcome the aforementioned problem by learning meta-weights that quickly adapt to visual concepts. However, MAML is heavily impacted by the meta-annotation phase for all images in the medical dataset \cite{nguyen2019overcoming}. Different from normal images, transfer learning in medical images is more challenging due to: \textit{(i)} noisy labels may occur when labeling images in an unsupervised manner;
\textit{(ii)} high-level semantic labels cause uncertainty during learning; and \textit{(iii)} difficulty in scaling up the process to all unlabeled images in medical datasets. 

In this paper, we introduce a new Multiple Meta-model Quantifying (MMQ) process to address these aforementioned problems in MAML. Intuitively MMQ is designed to: \textit{(i)} effectively increase meta-data by auto-annotation; 
\textit{(ii)} deal with the noisy labels in the training phase by leveraging the uncertainty of predicted scores during the meta-agnostic process; and \textit{(iii)} output meta-models which contain robust features for down-stream medical VQA task. Note that, compared with the recent approach for meta-learning in medical VQA \cite{nguyen2019overcoming}, our proposed MMQ does not take advantage of additional out-of-dataset images, while achieves superior accuracy in two challenging medical VQA datasets.

\section{Literature Review}
\label{Sec:Literature}

\textbf{Medical Visual Question Answering}
Based on the development of VQA in general images, 
the medical VQA task inherits similar techniques and achieves certain achievements~\cite{abacha2019vqa,lau2018dataset,Peng2018UMass,Abacha2018NML,zhou2018InceptionResnet,liu2020shengyan}. Specifically, the attention mechanisms such as MCB \cite{fukui2016multimodal}, SAN \cite{Yang2016StackedAN}, BAN \cite{Kim2018BilinearAN}, or CTI \cite{do2019compact} are applied in~\cite{Peng2018UMass,Abacha2018NML,zhou2018InceptionResnet,nguyen2019overcoming,vu2019ensemble} to learn joint representation between medical visual information and questions. Additionally, in \cite{lau2018dataset,zhou2018InceptionResnet,Peng2018UMass,kornuta2019leveraging}, the authors take advantage of transfer learning for extracting medical image features. Recently, approaches which directly solve different aspects of medical VQA are introduced, including reasoning \cite{kornuta2019leveraging,zhan2020medical}, diagnose model behavior \cite{vu2020questionCentric}, multi-modal fusion \cite{shi2019deep}, dedicated framework design \cite{lubna2019mobvqa,gupta2021hierarchical}, and generative model for dealing with abnormality questions \cite{ren2020cgmvqa}.

\textbf{Meta-learning}
Traditional machine learning algorithms, specifically deep learning-based approaches, require a large-scale labeled training set \cite{maicas2018training,nguyen2020autonomous,BarDWG15,nguyen2019scene,chi2020collaborative}. Therefore, meta-learning \cite{wang2016learning,schmidhuber1987evolutionary,HsuLF19,KhodadadehBS19}, which targets to deal with the problem of data limitation when learning new tasks, is applied broadly. 
There are three common approaches to meta-learning, namely model-based \cite{santoro2016meta,munkhdalai2017meta}, metric-based \cite{koch2015siamese,vinyals2016matching,sung2018learning,snell2017prototypical}, and optimization-based \cite{Ravi2017OptimizationAA,finn2017model,nichol2018first}. A notable optimization-based work, MAML \cite{finn2017model}, helps to learn a meta-model then quickly adapt it to other tasks. The authors in \cite{nguyen2019overcoming} used MAML to overcome the data limitation problem in medical VQA. However, their work required the use of external data during the training.

\section{Methodology}
\label{Sec:Method}
\subsection{Method overview}
\begin{figure}[!h]
  \centering
  \includegraphics[width=1.0\linewidth, height=0.5\linewidth]{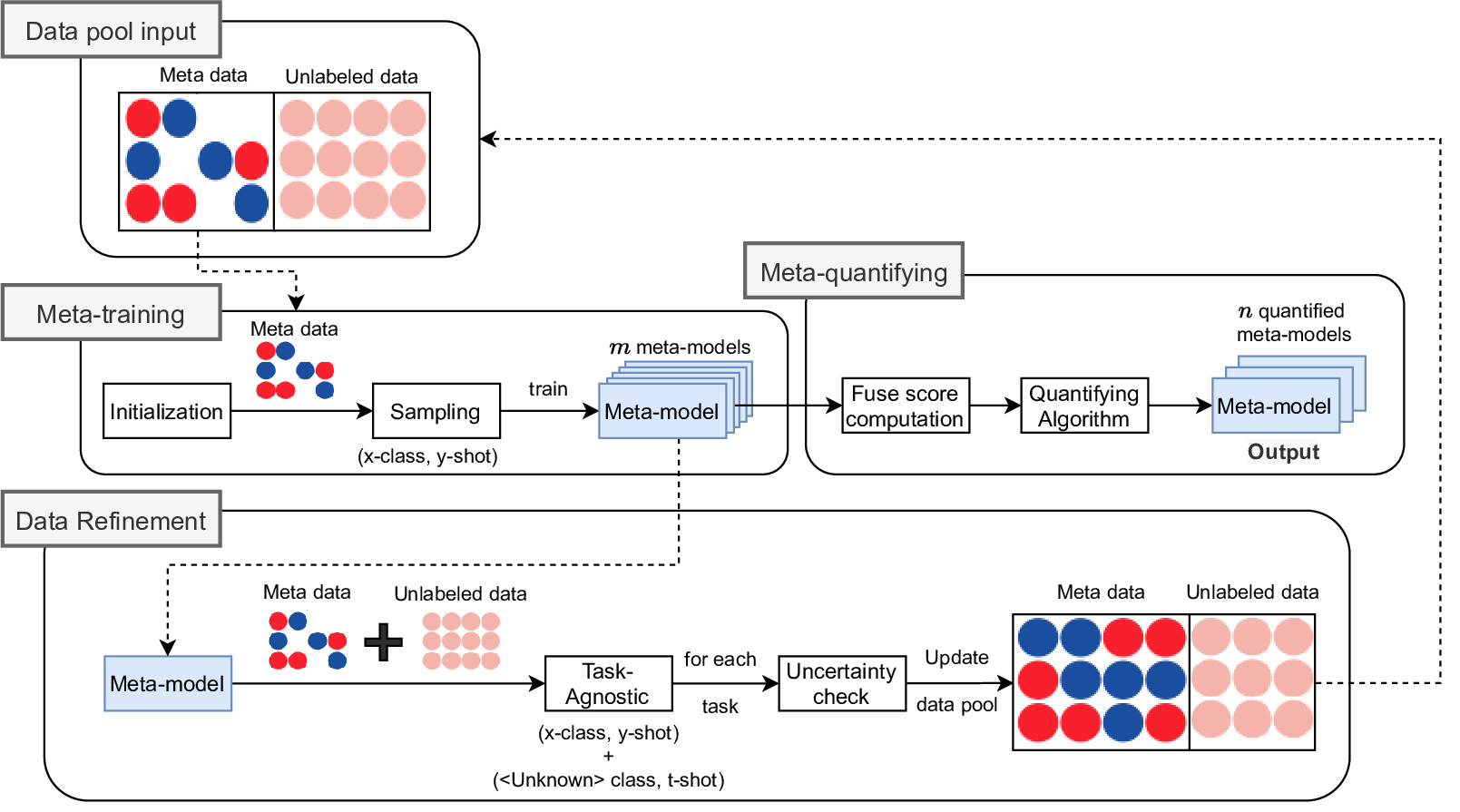}
 \caption{Multiple Meta-model Quantifying in medical VQA. Dotted lines denote looping steps, the number of loop equals to $m$ required meta-models.}
 \label{fig:proposal}
\end{figure}

\begin{figure}[!h]
  \centering
  \includegraphics[width=1.0\linewidth,height=0.4\linewidth]{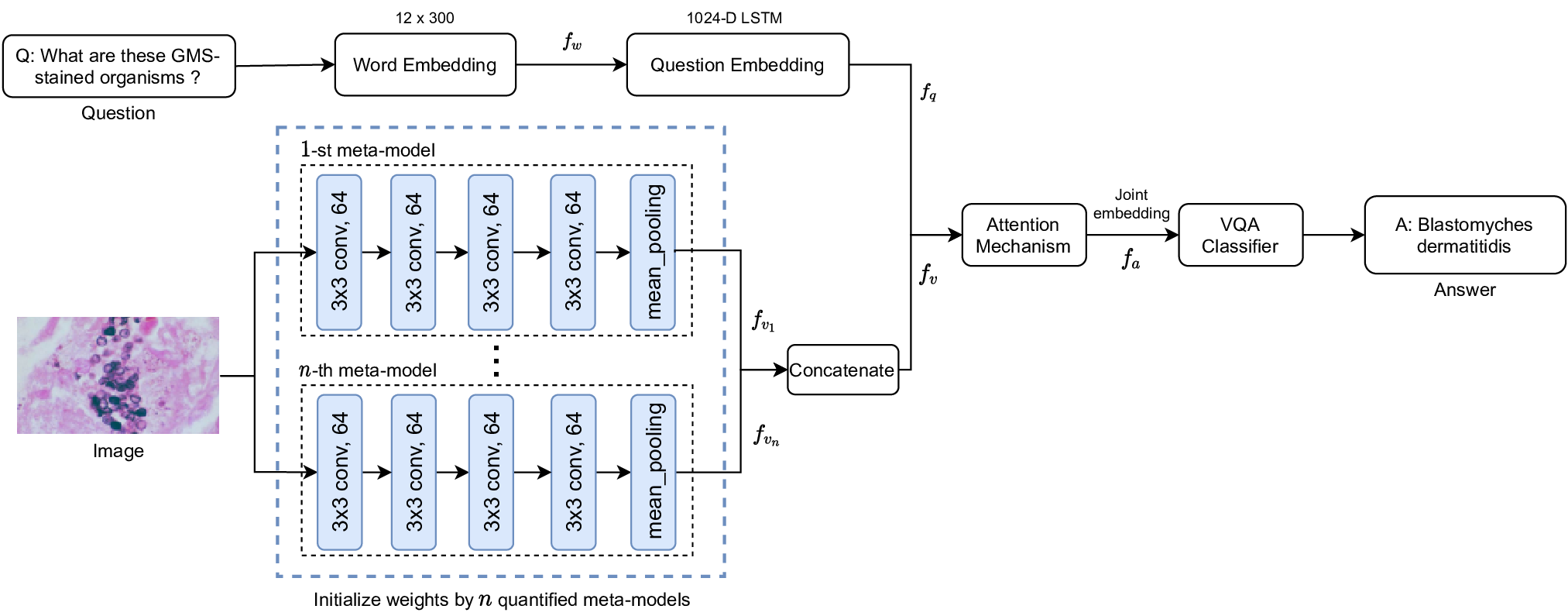}
 \caption{Our VQA framework is designed to integrate robust image features extracted from multiple meta-models outputted from MMQ.}
 \label{fig:VQAFramework}
\end{figure}

Our approach comprises two parts: our proposed multiple meta-model quantifying (MMQ - Figure \ref{fig:proposal}) and a VQA framework for integrating meta-models outputted from MMQ (Figure \ref{fig:VQAFramework}). MMQ addresses the meta-annotation problem by outputting multiple meta-models. These models are expected to robust to each other and have high accuracy during the inference phase of model-agnostic tasks. The VQA framework aims to leverage different features extracted from candidate meta-models and then generates predicted answers. 

\subsection{Multiple meta-model quantifying}
\label{subsec:gen_datapool_n_meta_model}
Multiple meta-model quantifying (Figure \ref{fig:proposal}) contains three modules: \textit{(i)} \textbf{Meta-training} which trains a specific meta-model for extracting image features used in medical VQA task by following MAML~\cite{finn2017model}; \textit{(ii)} \textbf{Data refinement} which increases the training data by auto-annotation and deal with the noisy label by leveraging the uncertainty of predicted scores; and \textit{(iii)} \textbf{Meta-quantifying} which selects meta-models whose robust to each others and have high accuracy during inference phase of model-agnostic tasks.

\SetKwInput{KwInput}{Input}                
\SetKwInput{KwOutput}{Output}              
\begin{algorithm}
\setcounter{AlgoLine}{0}
\DontPrintSemicolon
  \KwInput{$\rho(\mathcal{T})$ distribution over tasks; data pool $\mathcal{D}$; meta-model weights $\theta$}
  \KwOutput{Updated data pool $\mathcal{D}'$}
  Sample batch of tasks $\mathcal{T}_i \sim \rho(\mathcal{T})$
  
  Establish list $A$ with contains list of (Score $S$, Label $L$) of each sample in data pool $D$. $(S, L)$ is from the predicted process of Classifier $\mathcal{C}$ of each task $\mathcal{T}_i$.
  
  Set $\alpha$ and $\beta$ be uncertainty checking threshold.
  
  \textbf{For all} task $\mathcal{T}_i$ \textbf{do}
  
  \Indp \textbf{For all} image $I_k$ \textbf{in} $\mathcal{T}_i$ batch \textbf{do}
  
  \Indp ($S^i_k$, $L^i_k) \leftarrow$ $\mathcal{C}_i$($\mathcal{T}_i$, $\theta$, $I_k$). Where $\mathcal{C}_i$ is the $i$-th classifier of task  $\mathcal{T}_i$.
  
  Append ($S^i_k$, $L^i_k)$ into $A[I_k]$.
  
  \Indm \Indm Establish new version of Meta data split $\mathcal{M'}$ and new version of Unlabeled data split $\mathcal{U'}$ of $\mathcal{D}$
  
  \textbf{For all} element $A[I_j]$ \textbf{in} list $A$ \textbf{do}
  
  \Indp \textbf{If} $A[I_j]$ \textbf{in} Meta data split $\mathcal{M}$ of $\mathcal{D}$
  
  \Indp  \textbf{If} $\exists A[I_j]\{S\} < \alpha$ \textbf{and} $A[I_j]\{L\}$ is $A[I_j]\{GT_j\}$. Where $GT_j$ is the ground-truth label of $I_j$.
  
  \Indp Append $(I_j, A[I_j]\{L\})$ into $\mathcal{U}'$
  
  Remove $(I_j, A[I_j]\{L\})$ from $\mathcal{M}$
  
  \Indm \Indm  \textbf{If} $A[I_j]$ \textbf{in} Unlabeled data split $\mathcal{U}$ of $\mathcal{D}$
  
  \Indp  \textbf{If} $\exists A[I_j]\{S\} > \beta$
  
  \Indp Append $argmax_{A[I_j]\{S\}}(I_j, A[I_j]\{L\})$ into $\mathcal{M}'$
  
  \Indm \Indm \Indm$\mathcal{U}_f$ = $\mathcal{U}$ - $\mathcal{M}'$ + $\mathcal{U}'$. Where $\mathcal{U}_f$  is the updated Unlabeled data split of $\mathcal{D}'$. 
  
  $\mathcal{M}_f$ = $\mathcal{M}$ - $\mathcal{U}'$ + $\mathcal{M}'$. Where $\mathcal{M}_f$  is the updated Meta data split of $\mathcal{D}'$.
  
  \textbf{return} $\mathcal{M}_f, \mathcal{U}_f$ of $\mathcal{D}'$
\caption{Model-Agnostic for data refinement}
\label{alg:robust_check}
\end{algorithm}

\textbf{Meta-training} 
We generally follow MAML~\cite{finn2017model} to do meta-training. Let $f_{\theta}$ be the classification meta-model. Hence, $\theta$ represents the parameters of $f_{\theta}$ while $\{\theta'_{0}, \theta'_{1},...\theta'_{x}\}$ is the adapting parameters list of classification models for $x$ given tasks $\mathcal{T}_i$ and their associated dataset $\{\mathcal{D}^{tr}_i, \mathcal{D}^{val}_i\}$. 
Specifically, for each iteration, $x$ tasks are sampled with $y$ examples of each task. Then we calculate the gradient descent $\nabla_\theta L_{\mathcal{T}_i}(f_\theta (\mathcal{D}^{tr}_i))$ of the classification loss $L_{\mathcal{T}_i}$ and update the corresponding adapting parameters as follow.
\newcommand{\Lagr}{\mathcal{L}}
\begin{equation}
    \theta'_i = \theta - \alpha\nabla_\theta L_{\mathcal{T}_i}(f_\theta (\mathcal{D}^{tr}_i))
    \label{eq:1}
\end{equation}
At the end of each iteration, the meta-model parameters $\theta$ are updated throughout validation sets of all tasks sampled to learn the generalized features as:
\begin{equation}
    \theta \gets \theta - \beta\nabla_\theta\sum_{\mathcal{T}_i} L_{\mathcal{T}_i}(f_{\theta'_{i}}(\mathcal{D}^{val}_i)) 
    \label{eq:2}
\end{equation}
Unlike MAML~\cite{finn2017model} where only one meta-model is selected, we develop the following refinement and meta-quantifying steps to select high-quality meta-models for transfer learning to the medical VQA framework later. 

\textbf{Data refinement}
After finishing the meta-training phase, the weights of the meta-models are used for refining the dataset. The module aims to expand the meta-data pool for meta-training and removes samples that are expected to be hard-to-learn or have noisy labels (See Algorithm \ref{alg:robust_check} for more details). 

\textbf{Meta-quantifying}
\label{subsec:meta_quantify}
This module aims to identify candidate meta-models that are useful for the medical VQA task. A candidate model $\theta$ should achieve high performance during the validating process and its features distinct from other features from other candidate models.

To achieve these goals, we design a fuse score $S_F$ as described in (\ref{eq:F_score}). 
\begin{equation}
    S_F = \gamma S_P  + (1-\gamma)\sum_{t=1}^m 1 - Cosine \left(F_{c}, F_t\right) \forall F_{c} \neq  F_t
    \label{eq:F_score}
\end{equation}
where $S_P$ is the predicted score of the current meta-model over ground-truth label; $F_{c}$ is the feature extracted from the aforementioned meta-model that needs to compute the score; $F_t$ is the feature extracted from $t$-th model of the list of meta-model $\Theta$; Cosine is using for similarity checking between two features.

Since the predicted score $S_P$ at the ground-truth label and diverse score are co-variables, therefore the fuse score $S_F$ is also covariate with both aforementioned scores. This means that the larger $S_F$ is, the higher chance of the model to be selected for the VQA task. Algorithm \ref{alg:quantifying_meta} describes our meta-quantifying algorithm in details.
\SetKwInput{KwInput}{Input}                
\SetKwInput{KwOutput}{Output}              
\begin{algorithm}
\setcounter{AlgoLine}{0}
\DontPrintSemicolon
  \KwInput{Data pool $\mathcal{D}_T$; list of meta-model $\Theta \in [\theta_0, \theta_1,..., \theta_m]$ where $m$ denotes the number of candidate meta-models; number of quantified model $n$.}
  \KwOutput{List of Quantified meta-models $\Theta_{n} \in [\theta_0, \theta_1,..., \theta_n]$. $n < m$.}
  
  For all $n$ meta-models, sample batch of tasks $\mathcal{T}_i \sim \rho(\mathcal{T})$
  
  Establish list $A$ with contains list of (Score $S_P$, Feature $F)$ of each sample in quantify data pool $\mathcal{D}_T$. $(S_P,F)$ is got from the predicted process of Classifier $\mathcal{C}$ of each task $\mathcal{T}_i$. $S_P$ is the predicted score at ground-truth label.
  
  Set $\gamma$ be effectiveness - robustness balancing hyper-parameter.
  
  Establish Fuse Score list $\mathcal{L}_{S_F}$ for all meta-model in $\Theta$.
  
  \textbf{For all} task $\mathcal{T}_i$ \textbf{do}
  
  \Indp \textbf{For all} image $I_k$ \textbf{in} $\mathcal{T}_i$ batch \textbf{do}
  
   \Indp \textbf{For all} meta-model $\Theta_t$ \textbf{in} $\Theta$ \textbf{do}
  
  \Indp ($S^i_k$, $F^i_k)^{\Theta_t} \leftarrow$ $\mathcal{C}_i$($\mathcal{T}_i$, $\theta$, $I_k$). Where $\mathcal{C}_i$ is the $i$-th classifier of task  $\mathcal{T}_i$.
  
  Append ($S^i_k$, $F^i_k)^{\Theta_t}$ into $A[I_k]$.
  
   \Indm \textbf{For all} meta-model $\Theta_t$ \textbf{in} $\Theta$ \textbf{do}
   
  \Indp \textbf{For all} $A[I_k]$ \textbf{do}
  
  \Indp $S_F^{\Theta_t} \leftarrow$ Compute fuse score using Equation (\ref{eq:F_score}).
  
    $\mathcal{L}_{S_F}^{\Theta_t} += S_{F}^{\Theta_t}$.
    
    \Indm \Indm \Indm \Indm $\mathcal{L}_{S_F} \leftarrow $ Sort $\mathcal{L}_{S_F}$ decreasingly along with corresponding $\theta$.
    
    \textbf{return} $\Theta_{n} \leftarrow n$-first meta-models selected from $\mathcal{L}_{S_F}$.
\caption{Meta-quantifying algorithm}
\label{alg:quantifying_meta}
\end{algorithm}
\subsection{Integrate quantified meta-models to medical VQA framework}
To leverage robust features extracted from quantified meta-models, we introduce a VQA framework as in Figure \ref{fig:VQAFramework}.
Specifically, each input question is trimmed to a $12$-word sentence and then zero-padded if its length is less than $12$. Each word is represented by a 300-D GloVe word embedding \cite{pennington2014glove}. The word embedding is fed into a 1024-D LSTM to produce the question embedding $f_q$.

Each input image is passed through $n$ quantified meta-models got from the meta-quantifying module, which produce $n$ vectors. These vectors are concatenated to form an enhanced image feature, denoted as $f_v$ in Figure~\ref{fig:VQAFramework}. Since this vector contains multiple features extracted from different high-performed meta-models and each model has different views, the VQA framework is expected to be less affected by the bias problem. Image feature $f_v$ and question embedding $f_q$ are fed into an attention mechanism (BAN~\cite{Kim2018BilinearAN} or SAN~\cite{Yang2016StackedAN}) to produce a joint representation  $f_a$. This feature $f_a$ is used as input for a multi-class classifier (over the set of predefined answer classes~\cite{lau2018dataset}). To train the proposed model, we use a Cross Entropy loss for the answer classification task.
The whole VQA framework is then fine-tuned in an end-to-end manner.

\section{Experiments}
\label{Sec:Exp}
\subsection{Dataset}
\label{Sec:Dataset}
We use the VQA-RAD~\cite{lau2018dataset} and PathVQA \cite{he2020pathvqa} in our experiments.
The VQA-RAD~\cite{lau2018dataset} dataset contains 315 images and 3,515 corresponding questions. Each image is associated with more than one question.
The PathVQA \cite{he2020pathvqa} dataset consists of 32,799 question-answer pairs generated from 1,670 pathology images collected from
two pathology textbooks, and 3,328 pathology images collected from the PEIR digital library.

\subsection{Experimental details}
\label{Sec:ExpDetail}
\textbf{Meta-training.} Similar to~\cite{nguyen2019overcoming}, we first create the meta-annotation for training MAML. For the VQA-RAD dataset, we re-use the meta-annotation created by ~\cite{nguyen2019overcoming}. Note that we do not use their extra collected data in our experiment. For the PathVQA dataset, we create the meta-annotation by categorizing all training images into $31$ classes based on body parts, types of images, and organs.

For every iteration of MAML training, 
$5$ tasks are sampled per iteration in RAD-VQA while in PathVQA, this value is $4$ instead. 
For each task, in RAD-VQA, we randomly select $3$ classes from $9$ classes while in PathVQA we select $5$ classes from $31$ aforementioned classes. 
For each class, in RAD-VQA, we randomly select $6$ images in which $3$ images are used for updating task models and the remaining $3$ images are used for updating meta-model. In PathVQA, the same process is applied with $20$ random selected images, $5$ of them are used for updating task models and the remains are used for updating meta-model.

\textbf{Data refinement.} The meta-model outputted from the meta-training step is then used for updating the data pool through the algorithm described in Section \ref{subsec:gen_datapool_n_meta_model}. The refined data pool is then leveraged as the input for the meta-training step to output another meta-model. This loop is applied by a maximum of $7$ times to output up to $7$ different meta-models.

\textbf{Meta-quantifying.} All meta-models got from the previous step are passed through the Algorithm \ref{alg:quantifying_meta} to quantify their effectiveness. A maximum of $4$ models which have high performance is applied to VQA training.  

\textbf{VQA training.} 
After selecting candidate meta-models from the meta quantifying module, we use their trained weights to initialize the image feature extraction component in the VQA framework. We then finetune the whole VQA model using the VQA training set. The output vector of each meta-model is set to $32$-D in PathVQA and $64$-D in VQA-RAD dataset. We use $50\%$ of meta-annotated images for training meta-models. The effect of meta-annotated images can be found in our supplementary material.

\textbf{Baselines.} 
We compare our MMQ results with recent methods in medical VQA: MAML \cite{finn2017model}, MEVF \cite{nguyen2019overcoming}, stacked attention network (StAN) with VGG-16 \cite{he2020pathvqa}, and bilinear attention network (BiAN) with Faster-RCNN \cite{he2020pathvqa}. Two attentions methods SAN~\cite{Yang2016StackedAN} and BAN~\cite{Kim2018BilinearAN} are used in MAML, MEVF, and MMQ. Note that, StAN and BiAN only use pretrained models from the ImageNet dataset, MEVF \cite{nguyen2019overcoming} uses extra collected data to train their meta-model, while our MMQ relies solely on the images from the dataset. For the question feature extraction, all baselines and our method use the same pretrained models (i.e., Glove~\cite{pennington2014glove}) and then finetuning on VQA-RAD or PathVQA dataset.

\subsection{Results}
\label{Sec:Results}
Table~\ref{tab:sota} presents comparative results between different methods. The results show that our MMQ significantly outperforms other meta-learning methods by a large margin. Besides, the gain in performance of MMQ is stable with different attention mechanisms (BAN \cite{Kim2018BilinearAN} or SAN \cite{Yang2016StackedAN}) in the VQA task.
It worth noting that, compared with the most recent state-of-the-art method MEVF \cite{nguyen2019overcoming}, we outperform $5.3\%$ in free-form questions of the PathVQA dataset and $9.8\%$ in the Open-ended questions of the VQA-RAD dataset, respectively. 
Moreover, no out-of-dataset images are used in MMQ for learning meta-models.
The results imply that our proposed MMQ learns essential representative information from the input images and leverage effectively the features from meta-models to deal with challenging questions in medical VQA datasets.

\begin{table}[!t]
\centering
\setlength{\tabcolsep}{0.45 em} 
{\renewcommand{\arraystretch}{1.2}
\caption{Performance comparison on VQA-RAD and Path-VQA test set. (*) indicates methods used pre-trained model on ImageNet dataset. We refine data $5$ times ($m=5$) and use $3$ meta-models ($n=3$) in our MMQ.
}
\begin{tabular}{|c|c|c|c|c|c|c|c|}
\hline
\multirow{3}{*}{\textbf{\begin{tabular}[c]{@{}c@{}}Reference\\ Methods\end{tabular}}} & \multirow{3}{*}{\textbf{\begin{tabular}[c]{@{}c@{}}Attention\\ Method\end{tabular}}} & \multicolumn{3}{c|}{\textbf{PathVQA}}                                                                                                                                    & \multicolumn{3}{c|}{\textbf{VQA-RAD}}                                                                                                                                          \\ \cline{3-8} 
                                                                                                &                                                                                             & \textit{\textbf{\begin{tabular}[c]{@{}c@{}}Free-\\ form\end{tabular}}} & \textit{\textbf{\begin{tabular}[c]{@{}c@{}}Yes/\\ No\end{tabular}}} & \textit{\textbf{\begin{tabular}[c]{@{}c@{}}Over-\\ all\end{tabular}}} & \textit{\textbf{\begin{tabular}[c]{@{}c@{}}Open-\\ ended\end{tabular}}} & \textit{\textbf{\begin{tabular}[c]{@{}c@{}}Close-\\ ended\end{tabular}}} & \textit{\textbf{\begin{tabular}[c]{@{}c@{}}Over-\\ all\end{tabular}}} \\ \hline
\multirow{1}{*}{\begin{tabular}[c]{@{}c@{}}StAN~\cite{he2020pathvqa}(*)\end{tabular}}                & SAN                                                                                         & 1.6                                                                    & 59.4                                                                & 30.5                      & 24.2                                                                    & 57.2                                                                     & 44.2                      \\ \hline 
\multirow{1}{*}{\begin{tabular}[c]{@{}c@{}}BiAN~\cite{he2020pathvqa}(*)\end{tabular}}                & BAN                                                                                         & 2.9                                                                    & 68.2                                                                & 35.6                      & 28.4                                                                    & 67.9                                                                     & 52.3                      \\ \hline
                                  
\multirow{2}{*}{MAML \cite{finn2017model} }                                                                           & SAN                                                                                         & 5.4                                                                    & 75.3                                                               & 40.5                     & 38.2                                                                    & 69.7                                                                     & 57.1                      \\ \cline{2-8} 
                                                                                                & BAN                                                                                         & 5.9                                                                   & 79.5                                                               &  42.9                     & 40.1                                                                        & 72.4                                                                         &  59.6                         \\ \hline
\multirow{2}{*}{MEVF \cite{nguyen2019overcoming}}                                                                           & SAN                                                                                         & 6.0                                                                  & 81.0                                                                & 43.6                     & 40.7                                                                    & 74.1                                                                     & 60.7                      \\ \cline{2-8} 
                                                                                                & BAN                                                                                         & 8.1                                                                   & 81.4                                                                &  44.8                     & 43.9                                                                    & 75.1                                                                     & 62.7                      \\ \hline
\multirow{2}{*}{\textbf{\begin{tabular}[c]{@{}c@{}}MMQ (ours)\end{tabular}}}                  & SAN                                                                                      & 11.2                                                                      & 82.7                                                                   & 47.1                 &  46.3                                                                       &  75.7                                                                        & 64.0                          \\ \cline{2-8} 
                                                                                                & BAN                                                                                         & \textbf{13.4 }                                                                      & \textbf{84.0}                                                                    & \textbf{48.8}                          & \textbf{53.7}                                                                       & \textbf{75.8}                                                                          & \textbf{67.0}                          \\ \hline
\end{tabular}
\label{tab:sota}
}
\end{table}

\subsection{Ablation study}
\label{subsec:Abl}
Table \ref{tab:abl} presents our MMQ accuracy in PathVQA dataset when applying $m$ times refining data and $n$ quantified meta-models. The results show that, by using only $1$ quantified meta-model outputted from our MMQ, we significantly outperform both MAML and MEVF baselines. 
This confirms the effectiveness of the proposed MMQ for dealing with the limitation of meta-annotation in medical VQA, i.e., noisy labels and scalability. Besides, leveraging more quantified meta-models also further improves the overall performance.

We note that the improvements of our MMQ are more significant on free-form questions over yes/no questions. 
This observation implies that the free-form questions/answers which are more challenging and need more information from input images benefits more from our proposed method.

Table \ref{tab:abl} also shows that increasing the number of refinement steps and the number of quantified meta-models can improve the overall result, but the gain is smaller after each loop. The training time also increases when the number of meta-models is set higher. However, our testing time and the total number of parameters are only slightly higher than MAML \cite{finn2017model} and MEVF \cite{nguyen2019overcoming}. Based on the empirical results, we recommend applying $5$ times refinement with a maximum of $3$ quantified meta-models to balance the trade-off between the accuracy performance and the computational cost.

\begin{table}[!t]
\centering
\setlength{\tabcolsep}{0.3 em} 
{\renewcommand{\arraystretch}{1.2}
\caption{The effectiveness of our MMQ under $m$ times refining data and $n$ quantified meta-models on PathVQA test set. BAN is used as the attention method.}

\begin{tabular}{|c|c|c|c|c|c|c|c|c|}
\hline
\textbf{Methods}                                                                   & \textbf{\begin{tabular}[c]{@{}c@{}}\textbf{m}\end{tabular}} & \textbf{\begin{tabular}[c]{@{}c@{}}\textbf{n}\end{tabular}} & \textbf{\begin{tabular}[c]{@{}c@{}}Free-\\form\end{tabular}} & \textbf{\begin{tabular}[c]{@{}c@{}}Yes/\\No\end{tabular}} & \textbf{\begin{tabular}[c]{@{}c@{}}Over-\\all\end{tabular}} & \textbf{\begin{tabular}[c]{@{}c@{}}Train time\\(hours)\end{tabular}} & \textbf{\begin{tabular}[c]{@{}c@{}}Test time\\(s/sample)\end{tabular}} & \textbf{\begin{tabular}[c]{@{}c@{}}\#Paras\\(M)\end{tabular}} \\ \hline
\multirow{1}{*}{\begin{tabular}[c]{@{}c@{}}MAML \cite{finn2017model}\end{tabular}} & \_                                                                        & \_                                                                             &  5.9                                                                       &  79.5                                                                   & 42.9    & 2.1 & 0.007  & 27.2\\ 
                                                                                          \hline
                                                                                \multirow{1}{*}{\begin{tabular}[c]{@{}c@{}}MEVF \cite{nguyen2019overcoming}\end{tabular}} & \_                                                                        & \_                                                                             & 8.1                                                                   & 81.4                                                                &  44.8    &2.5 &0.008 &27.9                                                                  \\ \hline 
 \multirow{4}{*}{\begin{tabular}[c]{@{}c@{}}\textbf{MMQ (ours)}\end{tabular}} & 3                                                                        & 1                                                                              & 10.1                                                                      & 82.1                                                                   & 46.2    & 5.8 & 0.008 & 27.8                                                                                \\ \cline{2-9} 
                                                                                          & 4                                                                        & 2                                                                              & 12.0                                                                      & 83.0                                                                   & 47.6    & 7.3 & 0.009 & 28.1                                                                  \\ \cline{2-9} 
                                                                                          & 5                                                                       & 3                                                                              &  13.4                                                                     & 84.0                                                                    & 48.8   &8.9 &0.010 &28.3                 \\ \cline{2-9} 
                                                                                          & 7                                                                       & 4                                                                              &  13.6                                                                     & 84.0                                                                    & 48.8   &12.1 &0.011 &28.5                                                   \\ \hline
\end{tabular}
\label{tab:abl}
}
\end{table}

\section{Conclusion}
\label{Sec:Conclusion}
In this paper, we proposed a new multiple meta-model quantifying method to effectively leverage meta-annotation and deal with noisy labels in the medical VQA task. The extensively experimental results show that our proposed method outperforms the recent state-of-the-art meta-learning based methods by a large margin in both PathVQA and VQA-RAD datasets. Our implementation and trained models will be released for reproducibility.

\bibliographystyle{splncs04}
\bibliography{paper66}

\begin{thebibliography}{10}
\providecommand{\url}[1]{\texttt{#1}}
\providecommand{\urlprefix}{URL }
\providecommand{\doi}[1]{https://doi.org/#1}

\bibitem{Abacha2018NML}
Abacha, A.B., Gayen, S., Lau, J.J., Rajaraman, S., Demner-Fushman, D.: {NLM} at
  {ImageCLEF} 2018 visual question answering in the medical domain. {CEUR}
  Workshop Proceedings (2018)

\bibitem{abacha2019vqa}
Abacha, A.B., Hasan, S.A., Datla, V.V., Liu, J., Demner-Fushman, D.,
  M{\"u}ller, H.: Vqa-med: Overview of the medical visual question answering
  task at imageclef 2019. In: CLEF (Working Notes) (2019)

\bibitem{BarDWG15}
Bar, Y., Diamant, I., Wolf, L., Greenspan, H.: Deep learning with non-medical
  training used for chest pathology identification. In: Medical Imaging:
  Computer-Aided Diagnosis (2015)

\bibitem{chi2020collaborative}
Chi, W., Dagnino, G., Kwok, T.M., Nguyen, A., Kundrat, D., Abdelaziz, E., Riga,
  C., Bicknell, C., Yang, G.Z.: Collaborative robot-assisted endovascular
  catheterization with generative adversarial imitation learning. In: ICRA
  (2020)

\bibitem{do2019compact}
Do, T., Do, T.T., Tran, H., Tjiputra, E., Tran, Q.D.: Compact trilinear
  interaction for visual question answering. In: ICCV (2019)

\bibitem{finn2017model}
Finn, C., Abbeel, P., Levine, S.: Model-agnostic meta-learning for fast
  adaptation of deep networks. In: ICML (2017)

\bibitem{fukui2016multimodal}
Fukui, A., Park, D.H., Yang, D., Rohrbach, A., Darrell, T., Rohrbach, M.:
  Multimodal compact bilinear pooling for visual question answering and visual
  grounding. In: EMNLP (2016)

\bibitem{gupta2021hierarchical}
Gupta, D., Suman, S., Ekbal, A.: Hierarchical deep multi-modal network for
  medical visual question answering. Expert Systems with Applications  (2021)

\bibitem{He2016ResNet}
He, K., Zhang, X., Ren, S., Sun, J.: Deep residual learning for image
  recognition. In: CVPR (2016)

\bibitem{he2020pathvqa}
He, X., Zhang, Y., Mou, L., Xing, E., Xie, P.: Pathvqa: 30000+ questions for
  medical visual question answering. arXiv preprint arXiv:2003.10286  (2020)

\bibitem{HsuLF19}
Hsu, K., Levine, S., Finn, C.: Unsupervised learning via meta-learning. In:
  ICLR (2019)

\bibitem{huang2020tracking}
Huang, B., Tsai, Y.Y., Cartucho, J., Vyas, K., Tuch, D., Giannarou, S., Elson,
  D.S.: Tracking and visualization of the sensing area for a tethered
  laparoscopic gamma probe. International Journal of Computer Assisted
  Radiology and Surgery  (2020)

\bibitem{baurudepth21}
Huang, B., Zheng, J.Q., Nguyen, A., Tuch, D., Vyas, K., Giannarou, S., Elson,
  D.: Self-supervised generative adversarial network for depth estimation in
  laparoscopic images. In: MICCAI (2021)

\bibitem{KhodadadehBS19}
Khodadadeh, S., B{\""{o}}l{\""{o}}ni, L., Shah, M.: Unsupervised meta-learning
  for few-shot image classification. In: NIPS (2019)

\bibitem{Kim2018BilinearAN}
Kim, J.H., Jun, J., Zhang, B.T.: Bilinear attention networks. In: NIPS (2018)

\bibitem{koch2015siamese}
Koch, G., Zemel, R., Salakhutdinov, R.: Siamese neural networks for one-shot
  image recognition. In: ICML Deep Learning Workshop (2015)

\bibitem{kornuta2019leveraging}
Kornuta, T., Rajan, D., Shivade, C., Asseman, A., Ozcan, A.S.: Leveraging
  medical visual question answering with supporting facts. arXiv:1905.12008
  (2019)

\bibitem{lau2018dataset}
Lau, J.J., Gayen, S., Abacha, A.B., Demner-Fushman, D.: A dataset of clinically
  generated visual questions and answers about radiology images. Nature  (2018)

\bibitem{liu2020shengyan}
Liu, S., Ding, H., Zhou, X.: Shengyan at vqa-med 2020: An encoder-decoder model
  for medical domain visual question answering task. CLEF (2020)

\bibitem{lubna2019mobvqa}
Lubna, A., Kalady, S., Lijiya, A.: Mobvqa: A modality based medical image
  visual question answering system. In: TENCON (2019)

\bibitem{maicas2018training}
Maicas, G., Bradley, A.P., Nascimento, J.C., Reid, I., Carneiro, G.: Training
  medical image analysis systems like radiologists. In: MICCAI (2018)

\bibitem{munkhdalai2017meta}
Munkhdalai, T., Yu, H.: Meta networks. In: ICML (2017)

\bibitem{nguyen2019scene}
Nguyen, A.: Scene understanding for autonomous manipulation with deep learning.
  arXiv preprint arXiv:1903.09761  (2019)

\bibitem{nguyen2020end}
Nguyen, A., Kundrat, D., Dagnino, G., Chi, W., Abdelaziz, E., Guo, Y., Ma, Y.,
  Kwok, T., Riga, C., Yang, G.Z.: End-to-end real-time catheter segmentation
  with optical flow-guided warping during endovascular intervention. In: ICRA
  (2020)

\bibitem{nguyen2020autonomous}
Nguyen, A., Nguyen, N., Tran, K., Tjiputra, E., Tran, Q.: Autonomous navigation
  in complex environments with deep multimodal fusion network. In: IROS (2020)

\bibitem{nguyen2019overcoming}
Nguyen, B.D., Do, T.T., Nguyen, B.X., Do, T., Tjiputra, E., Tran, Q.D.:
  Overcoming data limitation in medical visual question answering. In: MICCAI
  (2019)

\bibitem{nichol2018first}
Nichol, A., Achiam, J., Schulman, J.: On first-order meta-learning algorithms.
  arXiv preprint arXiv:1803.02999  (2018)

\bibitem{Peng2018UMass}
Peng, Y., Liu, F., Rosen, M.P.: Umass at imageclef medical visual question
  answering (med-vqa) 2018 task. {CEUR} Workshop Proceedings (2018)

\bibitem{pennington2014glove}
Pennington, J., Socher, R., Manning, C.D.: Glove: Global vectors for word
  representation. In: EMNLP (2014)

\bibitem{Ravi2017OptimizationAA}
Ravi, S., Larochelle, H.: Optimization as a model for few-shot learning. In:
  ICLR (2017)

\bibitem{ren2020cgmvqa}
Ren, F., Zhou, Y.: Cgmvqa: a new classification and generative model for
  medical visual question answering. IEEE Access  (2020)

\bibitem{russakovsky2015imagenet}
Russakovsky, O., Deng, J., Su, H., Krause, J., Satheesh, S., Ma, S., Huang, Z.,
  Karpathy, A., Khosla, A., Bernstein, M., et~al.: Imagenet large scale visual
  recognition challenge. IJCV  (2015)

\bibitem{santoro2016meta}
Santoro, A., Bartunov, S., Botvinick, M., Wierstra, D., Lillicrap, T.:
  Meta-learning with memory-augmented neural networks. In: ICML (2016)

\bibitem{schmidhuber1987evolutionary}
Schmidhuber, J.: Evolutionary Principles in Self-referential Learning. (1987)

\bibitem{shi2019deep}
Shi, L., Liu, F., Rosen, M.P.: Deep multimodal learning for medical visual
  question answering. In: CLEF (Working Notes) (2019)

\bibitem{Simonyan2015VGG}
Simonyan, K., Zisserman, A.: Very deep convolutional networks for large-scale
  image recognition. In: ICLR (2015)

\bibitem{snell2017prototypical}
Snell, J., Swersky, K., Zemel, R.S.: Prototypical networks for few-shot
  learning. In: NIPS (2017)

\bibitem{sung2018learning}
Sung, F., Yang, Y., Zhang, L., Xiang, T., Torr, P.H., Hospedales, T.M.:
  Learning to compare: Relation network for few-shot learning. In: CVPR (2018)

\bibitem{vinyals2016matching}
Vinyals, O., Blundell, C., Lillicrap, T., Kavukcuoglu, K., Wierstra, D.:
  Matching networks for one shot learning. In: NIPS (2016)

\bibitem{vu2020questionCentric}
{Vu}, M.H., {Löfstedt}, T., {Nyholm}, T., {Sznitman}, R.: A question-centric
  model for visual question answering in medical imaging. IEEE TMI  (2020)

\bibitem{vu2019ensemble}
Vu, M., Sznitman, R., Nyholm, T., L{\"o}fstedt, T.: Ensemble of streamlined
  bilinear visual question answering models for the imageclef 2019 challenge in
  the medical domain. In: Conference and Labs of the Evaluation Forum (2019)

\bibitem{wang2016learning}
Wang, Y.X., Hebert, M.: Learning from small sample sets by combining
  unsupervised meta-training with cnns. In: NIPS (2016)

\bibitem{Yang2016StackedAN}
Yang, Z., He, X., Gao, J., Deng, L., Smola, A.J.: Stacked attention networks
  for image question answering. In: CVPR (2016)

\bibitem{zhan2020medical}
Zhan, L.M., Liu, B., Fan, L., Chen, J., Wu, X.M.: Medical visual question
  answering via conditional reasoning. In: ACM International Conference on
  Multimedia (2020)

\bibitem{zhou2018InceptionResnet}
Zhou, Y., Kang, X., Ren, F.: Employing {Inception-Resnet-v2} and {Bi-LSTM} for
  medical domain visual question answering. {CEUR} Workshop Proceedings (2018)

\end{thebibliography}

\title{Supplementary material:\\Multiple Meta-model Quantifying\\for Medical Visual Question Answering}

\author{Tuong Do\inst{1}\and
Binh X. Nguyen\inst{1}\and
Erman Tjiputra\inst{1}\and
Minh Tran\inst{1}\and\\
Quang D. Tran\inst{1}\and
Anh Nguyen\inst{2}}

\institute{AIOZ, Singapore \\ \email{\{tuong.khanh-long.do,binh.xuan.nguyen,erman.tjiputra,\\minh.quang.tran,quang.tran\}@aioz.io} \and University of Liverpool, UK \\
\email{anh.nguyen@liverpool.ac.uk}}

\maketitle              
\begin{abstract}
In this supplementary material, we provide further analysis of MMQ to verify its effectiveness for the medical VQA task. In particular, we illustrate the network structure to extract features from meta-models. We clarify the general setup and the details of the meta-annotation step in the PathVQA dataset. We also analyze the effect of different amounts of meta-annotated images for training meta-models. The experiments are conducted using the PathVQA and VQA-RAD datasets.
\end{abstract}

\section{Network to extract features from meta-models}
\label{Sec:model_struc}
\begin{figure}[!h]
  \centering
    \includegraphics[width=\linewidth]{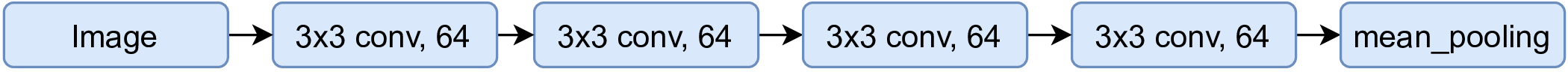}
 \caption{Feature extraction network from meta-models.}
 \label{fig:struct}
\end{figure}

Figure \ref{fig:struct} shows the network to extract features from the meta-model in our VQA framework. It consists of four $3\times3$ convolutional layers with stride $2$ and is ended with a mean pooling layer; each convolutional layer has $64$ filters and is followed by a ReLu layer. 
\section{General setup}
The image size is set at $84 \times 84$.
The proposed MMQ is implemented using PyTorch. The experiments are conducted on a single NVIDIA 1080Ti with 11GB RAM. In all MMQ experiment setups, $\alpha, \beta,$ and $\gamma$ are equalled to $0.01, 0.001,$ and $0.5$,  respectively. There is no fine-tuning process for these hyper-parameters since we use the default $\alpha, \beta$ values in MAML plus balanced initial $\gamma$ for training and still achieve good results.
\section{Meta-annotation details for PathVQA dataset}
For the PathVQA dataset, we create the meta-annotation by manually categorizing all training images into $31$ classes based on body parts, types of images, and organs.
These classes are: \textit{dense cell}, \textit{drawing}, \textit{process tree}, \textit{x-ray mouth}, \textit{x-ray ribs}, \textit{RGB bone}, \textit{RGB brain}, \textit{RGB endocrine}, \textit{RGB heart}, \textit{RGB intestine}, \textit{RGB kidney}, \textit{RGB liver}, \textit{RGB lung}, \textit{RGB mouth}, \textit{RGB skull}, \textit{RGB uterus}, \textit{RBG arms}, \textit{RGB baby}, \textit{RGB head}, \textit{RGB skin}, \textit{sparse cell}, \textit{chart}, \textit{x-ray legs}, \textit{RGB prostate}, \textit{RGB spleen}, \textit{RBG body}, \textit{RGB legs}, \textit{x-ray arms}, \textit{RBG oral}, \textit{RGB pancreas}, \textit{RGB penis}. 

These meta-labels, our source code and trained models will be release for reproducibility.

\section{Meta-data vs. Unlabeled data in MMQ}
Table \ref{tab:amount_meta_data} illustrates the performance of MMQ with different amounts of meta-annotated images for training meta models. Since our data refinement module (See Algorithm 1 in our paper) expands current meta-data as well as removes uncertainty samples simultaneously, keeping the balance in the number of samples between meta-data and unlabeled data at the initial step is worthy. Empirical results also imply that the initial data balance between two data pools greatly increases the effectiveness of our proposed  MMQ.

\begin{table}[!h]
\centering
\setlength{\tabcolsep}{0.45 em} 
{\renewcommand{\arraystretch}{1.2}
\caption{Performance (\%) comparison on VQA-RAD and Path-VQA test set when using MMQ with different amount of meta-annotated images for training meta-models. These results reported after $5$ times refining data and $3$ quantified meta-models are picked up.
}
\begin{tabular}{|c|c|c|c|c|c|c|c|}
\hline
\multirow{3}{*}{\textbf{\begin{tabular}[c]{@{}c@{}}\% annotated \\ data\end{tabular}}} & \multirow{3}{*}{\textbf{\begin{tabular}[c]{@{}c@{}}Attention\\ Mechanism\end{tabular}}} & \multicolumn{3}{c|}{\textbf{PathVQA}}                                                                                                                                    & \multicolumn{3}{c|}{\textbf{VQA-RAD}}                                                                                                                                          \\ \cline{3-8} 
                                                                                                &                                                                                             & \textit{\textbf{\begin{tabular}[c]{@{}c@{}}Free-\\ form\end{tabular}}} & \textit{\textbf{\begin{tabular}[c]{@{}c@{}}Yes/\\ No\end{tabular}}} & \textit{\textbf{\begin{tabular}[c]{@{}c@{}}Over-\\ all\end{tabular}}} & \textit{\textbf{\begin{tabular}[c]{@{}c@{}}Open-\\ ended\end{tabular}}} & \textit{\textbf{\begin{tabular}[c]{@{}c@{}}Close-\\ ended\end{tabular}}} & \textit{\textbf{\begin{tabular}[c]{@{}c@{}}Over-\\ all\end{tabular}}} \\ \hline
\multirow{2}{*}{\textbf{\begin{tabular}[c]{@{}c@{}}25 \%\end{tabular}}}                  & SAN                                                                                      &7.2	&82.8	&45.1	&44.7	&72.4	&61.4                          \\ \cline{2-8} 
                                                                                                & BAN                                                                                         &9.8	&83.1	&46.5	&48.8	&74.6	&64.3                          \\ \hline
\multirow{2}{*}{\textbf{\begin{tabular}[c]{@{}c@{}}50\%\end{tabular}}}                  & SAN                                                                                      & 11.2                                                                      & 82.7                                                                   & 47.1                 &  46.3                                                                       &  75.7                                                                        & 64.0                          \\ \cline{2-8} 
                                                                                                & BAN                                                                                         & 13.4                                                                       & 84.0                                                                    & 48.8                          & 53.7                                                                        & 75.8                                                                          & 67.0                          \\ \hline
\multirow{2}{*}{\textbf{\begin{tabular}[c]{@{}c@{}}75\%\end{tabular}}}                  & SAN                                                                                      &10.1	&82.7	&46.5	&46.3	&74.6	&63.3                       \\ \cline{2-8} 
                                                                                                & BAN      &12.9	&83.3	&48.2	&48.8	&78.8	&66.2                         \\ \hline
\multirow{2}{*}{\textbf{\begin{tabular}[c]{@{}c@{}}100\%\end{tabular}}}                  & SAN                                                                                     &10.6	&82.8	&46.8	&47.2	&74.6	&63.6                        \\ \cline{2-8} 
                                                                                                & BAN                                                                                        &13.6	&83.2	&48.5	&48	&78.9	&66.6                         \\ \hline
\end{tabular}
\label{tab:amount_meta_data}
}
\end{table}

\section{Visualization for data refinement}
\begin{figure}[]
  \centering
    \subfigure[]{\includegraphics[width=0.177\linewidth, height=0.79\linewidth]{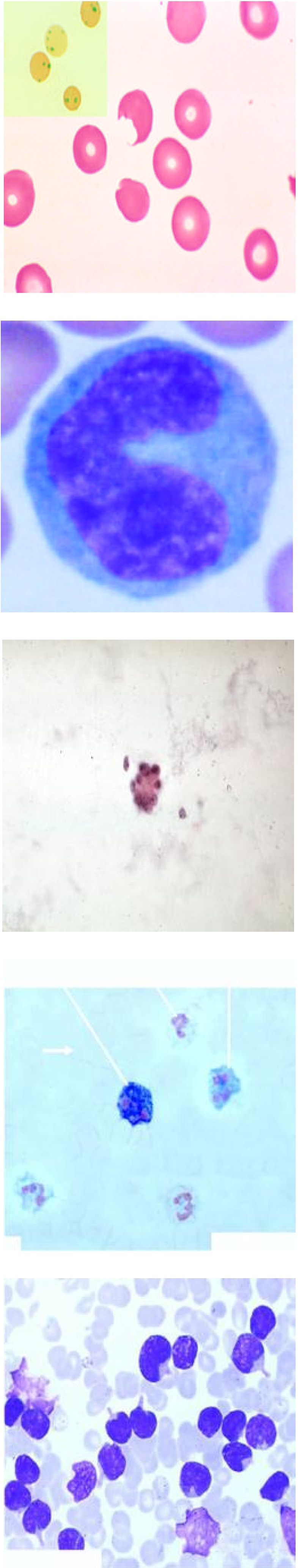}}
    \subfigure[]{\includegraphics[width=0.177\linewidth, height=0.79\linewidth]{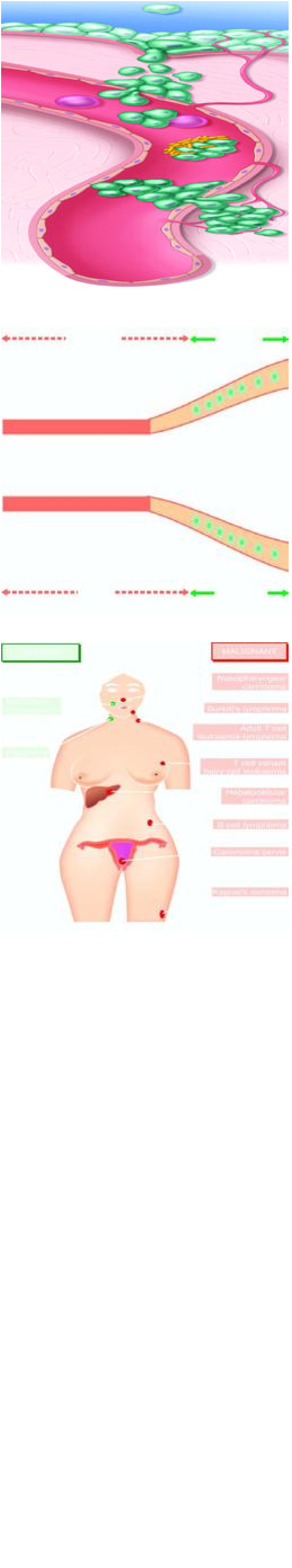}}
    \subfigure[]{\includegraphics[width=0.177\linewidth, height=0.79\linewidth]{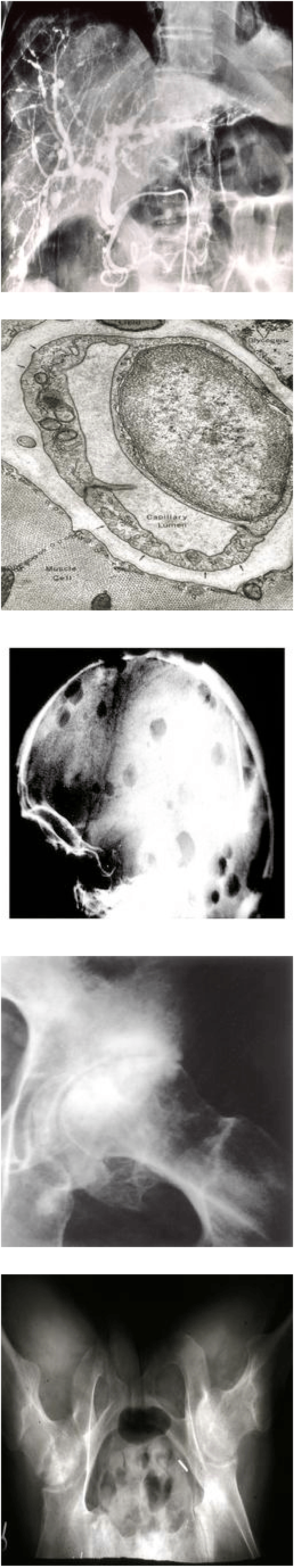}}
    \subfigure[]{\includegraphics[width=0.177\linewidth, height=0.79\linewidth]{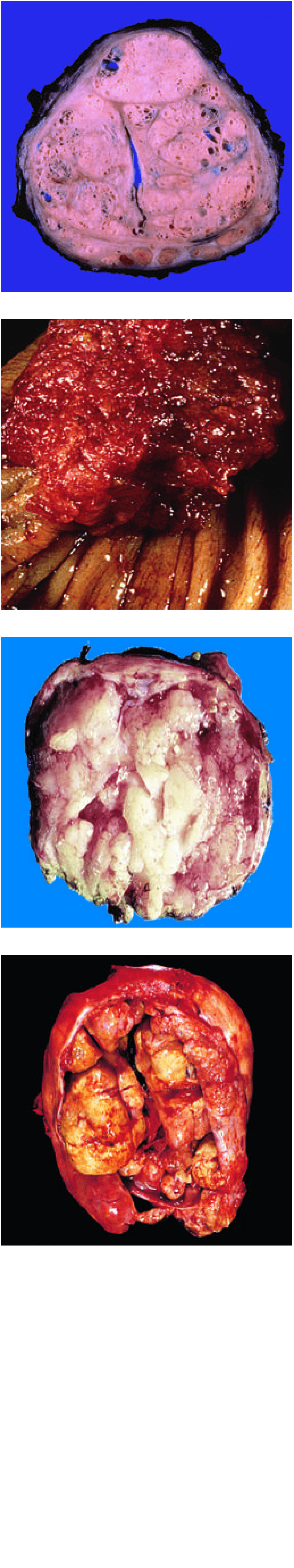}}
    \subfigure[]{\includegraphics[width=0.177\linewidth, height=0.79\linewidth]{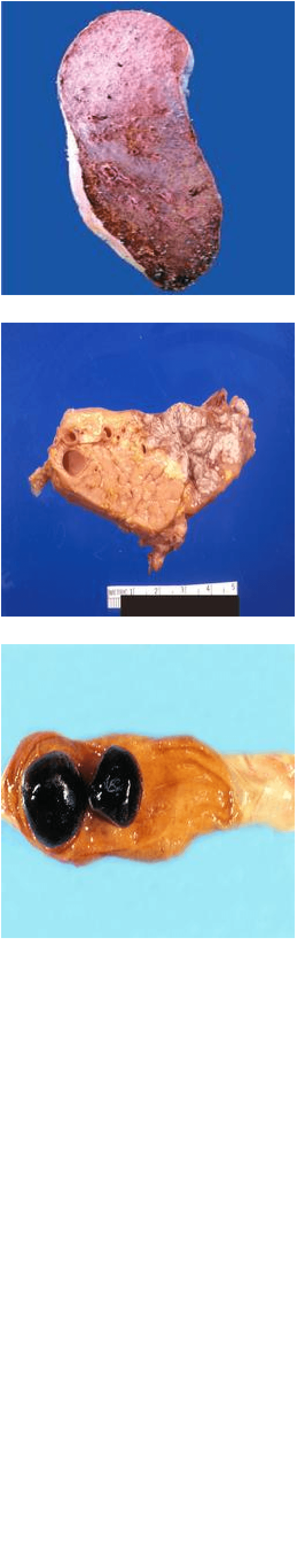}}
 \caption{The visualization images from meta-annotated data which are removed during the first refinement step, i.e., caused by their high uncertainty scores . Their labels are: \textbf{(a)} Dense Cell, \textbf{(b)} Process Tree, \textbf{(c)} X-ray Mouth, \textbf{(d)} RBG Brain, and \textbf{(e)} RBG Kidney, consequently. Best viewed in color.}
 \label{fig:remove}
\end{figure}

\begin{figure}[]
  \centering
    \subfigure[]{\includegraphics[width=0.177\linewidth, height=0.79\linewidth]{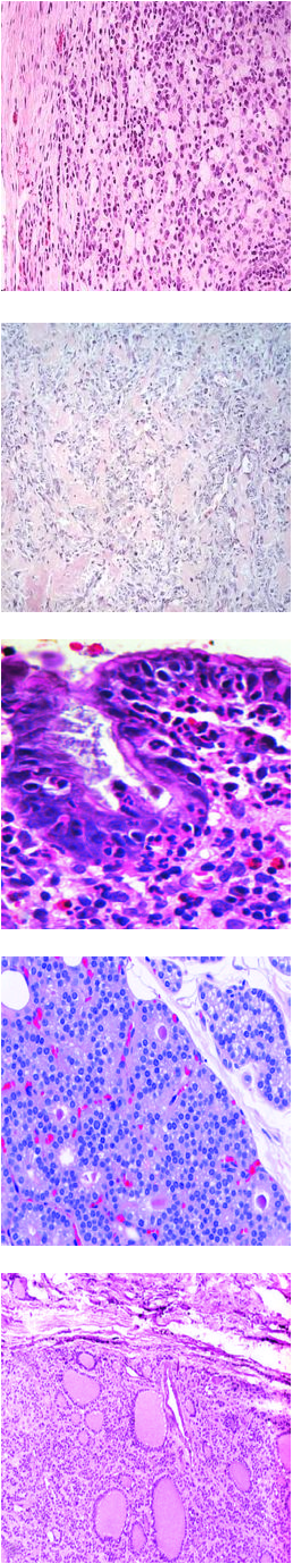}}
    \subfigure[]{\includegraphics[width=0.177\linewidth, height=0.79\linewidth]{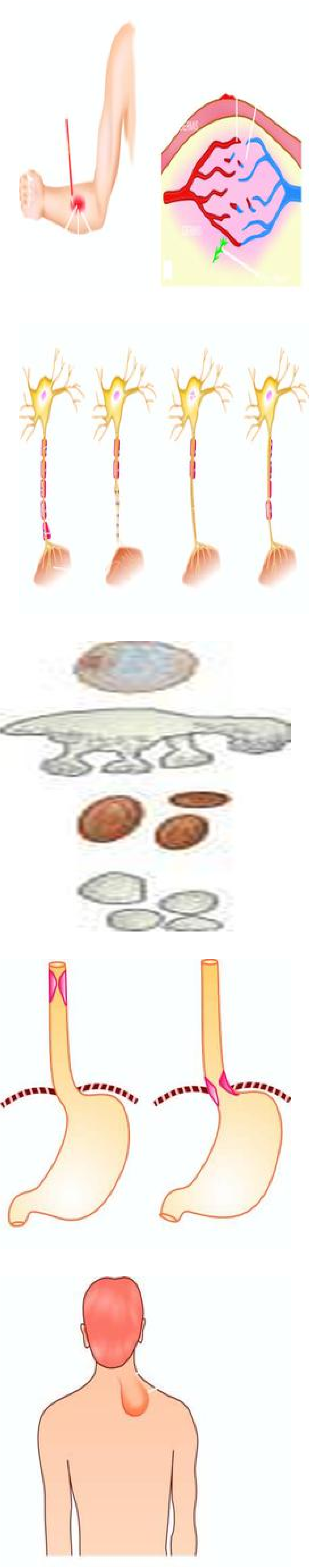}}
    \subfigure[]{\includegraphics[width=0.177\linewidth, height=0.79\linewidth]{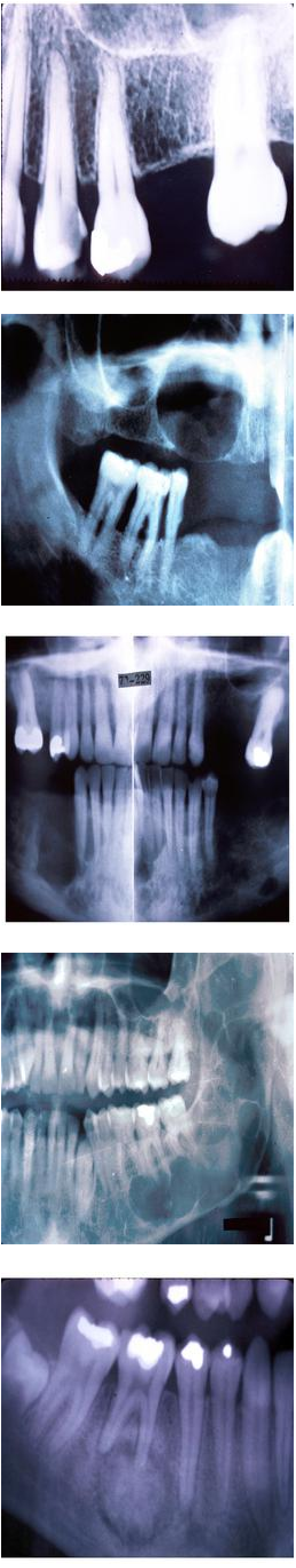}}
    \subfigure[]{\includegraphics[width=0.177\linewidth, height=0.79\linewidth]{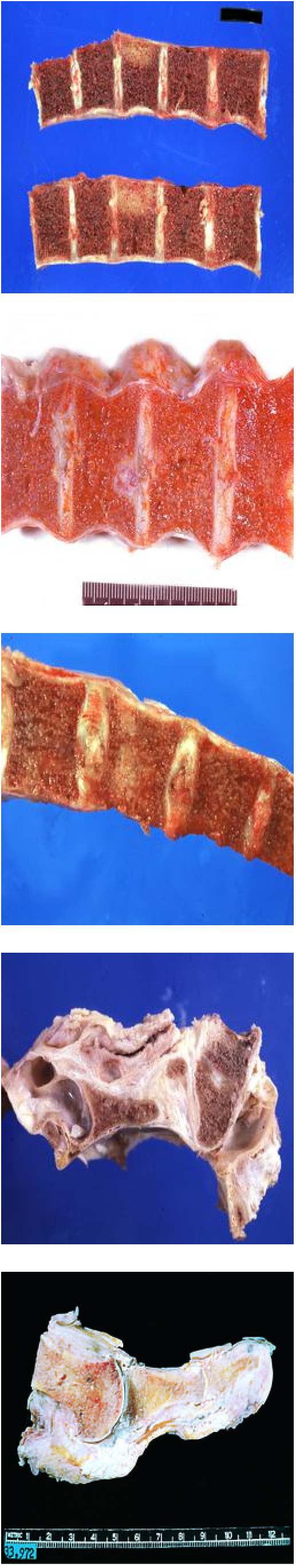}}
    \subfigure[]{\includegraphics[width=0.177\linewidth, height=0.79\linewidth]{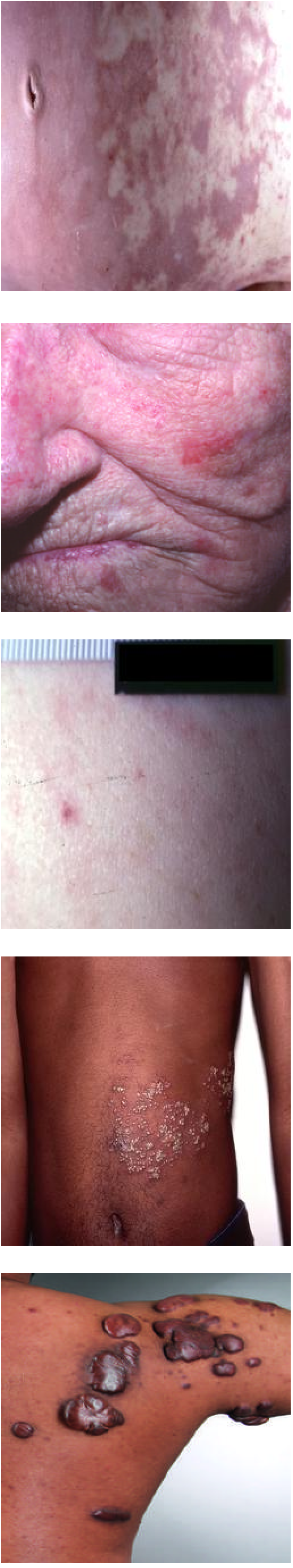}}
 \caption{The visualization images from unlabeled data which are chosen to add into meta-annotated data during the first refinement step. Their labels are: \textbf{(a)} Dense Cell, \textbf{(b)} Drawing, \textbf{(c)} X-ray Mouth, \textbf{(d)} RBG Bone, and \textbf{(e)} RBG Skin, consequently. Best viewed in color.}
 \label{fig:add}
\end{figure}

For visualization, Figure \ref{fig:remove} shows the images removed from the meta-annotated dataset due to their high uncertainty score. Additionally, Figure \ref{fig:add} illustrates the images and their labels which are annotated automatically by passing unlabeled image data through the $1$-st refinement step. These illustrations indicate that our MMQ successfully extend dataset by labelling meta-data automatically as well as remove samples with noisy labels. 

\end{document}